%% file: arxiv.tex
\documentclass[10pt,twocolumn,letterpaper]{article}

\usepackage{iccv}
\usepackage{times}
\usepackage{epsfig}
\usepackage{graphicx}
\usepackage{amsmath}
\usepackage{amssymb}

\usepackage{multirow}
\usepackage{algorithm}  
\usepackage{algorithmicx}  
\usepackage{algpseudocode}  
\usepackage{cite}
\usepackage{subfig}
\usepackage{array}
\usepackage{colortbl}
\usepackage[accsupp]{axessibility}

\usepackage[table]{xcolor}
\definecolor{col1}{RGB}{232, 161, 148}
\definecolor{col2}{RGB}{148, 187, 232}
\newcommand{\tabincell}[2]{\begin{tabular}{@{}#1@{}}#2\end{tabular}}

\usepackage[pagebackref=true,breaklinks=true,letterpaper=true,colorlinks,bookmarks=false]{hyperref}

\newcommand{\expo}[1]{\exp\left(#1\right)}

\newcommand{\absrp}{\sigma}

\newcommand{\timenear}{t_n}
\newcommand{\timefar}{t_f}
\newcommand{\deltatime}{\delta}
\iccvfinalcopy 

\def\iccvPaperID{4076} 

\ificcvfinal\fi

\begin{document}

\title{NerfingMVS: Guided Optimization of Neural Radiance Fields \\ for Indoor Multi-view Stereo}

\author{Yi Wei\textsuperscript{1,2}, Shaohui Liu\textsuperscript{3}, Yongming Rao\textsuperscript{1,2}, Wang Zhao\textsuperscript{4}, Jiwen Lu\textsuperscript{1,2}\thanks{Corresponding author}, Jie Zhou\textsuperscript{1,2}\\
\textsuperscript{1}Department of Automation, Tsinghua University, China\\
\textsuperscript{2}State Key Lab of Intelligent Technologies and Systems, China\\
\textsuperscript{3}ETH Zurich \quad
\textsuperscript{4}Department of Computer Science and Technology, Tsinghua University, China \\
\tt\small y-wei19@mails.tsinghua.edu.cn; b1ueber2y@gmail.com; raoyongming95@gmail.com; \\ \tt\small zhao-w19@mails.tsinghua.edu.cn; \{lujiwen, jzhou\}@tsinghua.edu.cn \\
}
\twocolumn[{%

\vspace{-1em}
\maketitle
\vspace{-1em}
\input{figures/teaser}
}] 
\ificcvfinal\thispagestyle{empty}\fi

\begin{abstract}

\input{latex/abstract.tex}
\end{abstract}

\vspace{-10pt}
\section{Introduction}
\input{latex/intro.tex}

\section{Related Work}
\input{latex/related.tex}

\section{Approach}
\input{latex/method.tex}

\section{Experiments}
\input{latex/exp.tex}

\section{Conclusion and Future Work}
\input{latex/conclusion.tex}

\noindent
\textbf{Acknowledgements: }
\input{latex/acknowledgement.tex}

\clearpage
{\small
\bibliographystyle{ieee_fullname}
\bibliography{egbib}
}

\clearpage
\twocolumn[{
	\vspace{-3em}
\input{tables/ablation-hyper-supp}
}]
\section*{Appendix}
\appendix
\input{latex/supp-arxiv}

\end{document}

%% file: figures/teaser.tex
\begin{center}
\centering
\vspace{-15pt}
\setlength\tabcolsep{1.0pt} 
\renewcommand{\arraystretch}{1.0}
\begin{tabular}{ccccc}
{\includegraphics[width=0.2\linewidth]{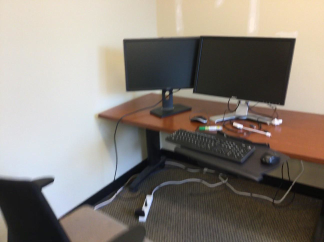}} & 
{\includegraphics[width=0.2\linewidth]{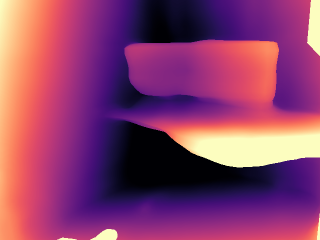}} &
{\includegraphics[width=0.2\linewidth]{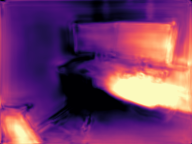}}&
{\includegraphics[width=0.2\linewidth]{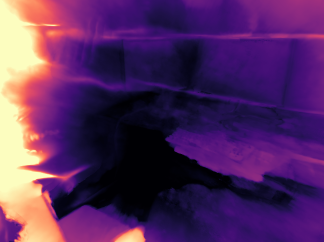}} & 
{\includegraphics[width=0.2\linewidth]{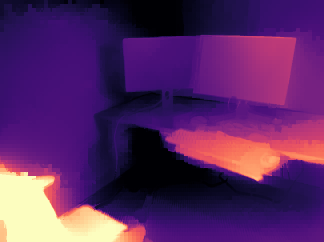}} \\
{\includegraphics[width=0.2\linewidth]{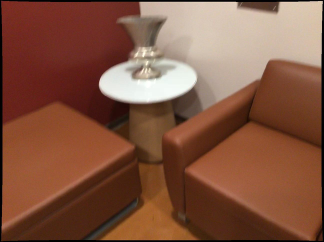}} & 
{\includegraphics[width=0.2\linewidth]{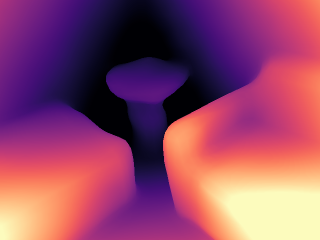}} &
{\includegraphics[width=0.2\linewidth]{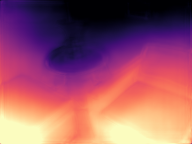}}&
{\includegraphics[width=0.2\linewidth]{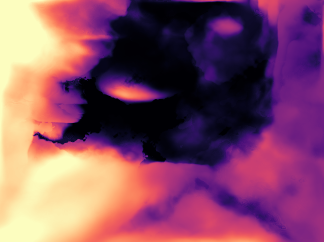}} & 
{\includegraphics[width=0.2\linewidth]{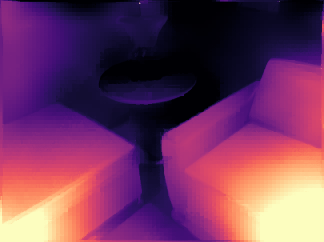}} \\
RGB & Atlas \cite{murez2020atlas} & CVD \cite{luo2020consistent} & NeRF \cite{mildenhall2020nerf} & Ours \\
\end{tabular}
\centering
\captionof{figure}{Qualitative results for multi-view depth estimation on ScanNet \cite{dai2017scannet}. Our method clearly surpasses leading multi-view estimation methods \cite{murez2020atlas,luo2020consistent} by building on top of neural radiance fields \cite{mildenhall2020nerf}. While also using test-time optimization, CVD \cite{luo2020consistent} suffers from inaccurate estimation of flow correspondences. NeRF \cite{mildenhall2020nerf} fails to produce accurate geometry due to the inherent shape-radiance ambiguity \cite{zhang2020nerf++} (See Figure \ref{fig::analysis}) in indoor scenes.With guided optimization, our method successfully integrates the learning-based depth priors into NeRF, significantly improving the geometry of the radiance fields. }
\label{fig::teaser}
\end{center}

%% file: latex/abstract.tex
In this work, we present a new multi-view depth estimation method that utilizes both conventional reconstruction and learning-based priors over the recently proposed neural radiance fields (NeRF). Unlike existing neural network based optimization method that relies on estimated correspondences, our method directly optimizes over implicit volumes, eliminating the challenging step of matching pixels in indoor scenes. The key to our approach is to utilize the learning-based priors to guide the optimization process of NeRF. Our system firstly adapts a monocular depth network over the target scene by finetuning on its sparse SfM+MVS reconstruction from COLMAP. Then, we show that the shape-radiance ambiguity of NeRF still exists in indoor environments and propose to address the issue by employing the adapted depth priors to monitor the sampling process of volume rendering. Finally, a per-pixel confidence map acquired by error computation on the rendered image can be used to further improve the depth quality. Experiments show that our proposed framework significantly outperforms state-of-the-art methods on indoor scenes, with surprising findings presented on the effectiveness of correspondence-based optimization and NeRF-based optimization over the adapted depth priors. In addition, we show that the guided optimization scheme does not sacrifice the original synthesis capability of neural radiance fields, improving the rendering quality on both seen and novel views. Code is available at \href{https://github.com/weiyithu/NerfingMVS}{\color{cyan}{https://github.com/weiyithu/NerfingMVS}}.

%% file: latex/intro.tex
Reconstructing 3D scenes from multi-view posed images, also named as multi-view stereo (MVS), has been a fundamental topic in computer vision over decades. The application varies from robotics, 3D modeling, to virtual reality, etc. Conventional multi-view stereo approaches \cite{yoon2006adaptive,gallup2007real,bleyer2011patchmatch,hosni2012fast} densely match pixels across views by comparing the similarity of cross-view image patches. While producing impressive results, those methods often suffer from poorly textured regions, thin structures and non-Lambertian surfaces, especially in real-world indoor environments.  

Recently, with the success of deep neural networks, several learning-based methods \cite{wang2018mvdepthnet,im2019dpsnet,liu2019neural,kusupati2020normal} are proposed to tackle the multi-view stereo problem often by employing a cost volume based architecture. Those methods perform a direct neural network inference at test time for multi-view depth estimation and achieve remarkable performance on benchmarks. However, due to the lack of constraints at inference, the predicted depth maps across views are often not consistent and the photometric consistency is often violated. To address this issue, \cite{luo2020consistent} proposed a test-time optimization framework that optimizes over learning-based priors acquired from single-image depth estimation. While being computationally inefficient, the method produces accurate and consistent depth maps that are available for various visual effects. However, the optimization formulation of this method relies heavily on an optical flow network \cite{ilg2017flownet} to establish correspondences, which becomes problematic when estimated correspondences are unreliable. 

In this paper, we present a new neural network based optimization framework for multi-view depth estimation based on the recently proposed neural radiance fields \cite{mildenhall2020nerf}. Instead of relying on estimated correspondences and cross-view depth reprojection for optimization \cite{luo2020consistent}, our method directly optimizes over volumes. However, we show that the shape-radiance ambiguity \cite{zhang2020nerf++} of NeRF becomes the bottleneck on estimating accurate per-view depths in indoor scenes. To address the issue, we propose a guided optimization scheme to help train NeRF with learning-based depth priors. Specifically, our system firstly adapts a monocular depth network onto the test scene by finetuning on its conventional SfM+MVS reconstruction. Then, we employ the adapted depth priors to guide the sampling process of volume rendering for NeRF. Finally, we acquire a confidence map from the rendered RGB image of NeRF and improve the depth map with a post-filtering step.

Our findings indicate that the scene-specific depth prior adaptation significantly improves the depth quality. However, performing existing correspondence-based optimization on the adapted depth priors will surprisingly degrade the performance. On the contrary, with direct optimization over neural radiance fields, our method consistently improves the depth quality over adapted depth priors. This phenomenon demonstrates the potential of exploiting neural radiance fields for accurate depth estimation. 

Experiments show that our proposed framework significantly improves upon state-of-the-art multi-view depth estimation methods on tested indoor scenes. In addition, the guided optimization from learning-based priors can help improve the rendering quality of NeRF on both seen and novel views, achieving comparable or better quality with state-of-the-art novel view synthesis methods. This indicates that conventional non-learning reconstruction method, while demonstrated to be effective on helping image-based view synthesis in \cite{riegler2020free,riegler2020stable}, can also help improve the synthesis quality on neural implicit representations.

%% file: latex/related.tex
\paragraph{Multi-view Reconstruction: }
Recently, 3D vision \cite{hou20193d,zhao2020towards,wei2019conditional,wei2021fgr,wei2021pv,tao2020seggroup}has attracted more and more attention.  Early multi-view reconstruction approaches include volumetric optimization \cite{kutulakos2000theory, faugeras2002variational,vogiatzis2005multi}, which perform global optimization with photo-consistency based assumptions. However, those methods suffer from large computational complexity. Another direction \cite{gallup2007real,bleyer2011patchmatch} is to estimate per-view depth map. Compared to volumetric approaches, these methods can produce finer geometry. However, they rely on accurately matched pixels by comparing the similarity of cross-view patches at different depth hypotheses, which will be problematic over poorly textured regions in indoor scenes. Recently, a number of learning-based methods are proposed. While some of them predict on voxelized grids \cite{kar2017learning, tatarchenko2017octree}, they suffer from limited resolution. An exception of this is Atlas \cite{murez2020atlas}, which predicts TSDF values via back-projection of the image features. Most learning-based methods \cite{wang2018mvdepthnet,im2019dpsnet,liu2019neural,hou2019multi,kusupati2020normal,long2020occlusion} follow the spirit of conventional approaches \cite{gallup2007real} and generate per-view depth map from a cost volume based architecture. Most related to us, \cite{luo2020consistent} performs test-time optimization over per-view depth maps with learning-based priors. While our work also utilizes the learning-based priors, we build on top of the recently proposed neural radiance fields \cite{martin2020nerf} and introduce a new way to accurately estimate multi-view depths by directly optimizing over implicit volumes with the guide of learning-based priors. Our method neither suffers from the resolution problem nor relies on accurately estimated correspondences. 
\input{figures/pipeline}
\vspace{-10pt}
\paragraph{Neural Implicit Representation:}
Recently, several seminal works \cite{mescheder2019occupancy,park2019deepsdf,chen2019learning} demonstrate the potential of representing implicit surfaces with a neural network, which enables memory-efficient geometric representation with infinite resolution. Variations include applying neural implicit representations on part hierarchies \cite{jiang2020local,genova2020local}, human reconstruction \cite{saito2019pifu,saito2020pifuhd}, view synthesis \cite{lombardi2019neural,sitzmann2019scene}, differentiable rendering \cite{niemeyer2020differentiable,liu2020dist}, etc. Neural radiance fields (NeRF) \cite{mildenhall2020nerf} represent scenes as continuous implicit function of positions and orientations for high quality view synthesis, which leads to several follow-up works \cite{zhang2020nerf++,bi2020neural,rebain2020derf} improving its performance. There are several extensions for NeRF including dynamic scenes \cite{park2020deformable,xian2020space}, portrait avatars \cite{gao2020portrait}, relighting \cite{bi2020neural}, pose estimation \cite{yen2020inerf}, etc. In this paper, we propose a guided optimization scheme to enrich NeRF \cite{mildenhall2020nerf} with the ability of accurate depth estimation which surpasses leading multi-view depth estimation approaches.

\vspace{-10pt}
\paragraph{View Synthesis:}
View synthesis is conventionally often referred as view interpolation \cite{gortler1996lumigraph, levoy1996light}, where the goal is to interpolate views within the convex hull of the initial camera positions. With the success of deep learning, learning-based methods \cite{zhou2018stereo, srinivasan2019pushing,flynn2019deepview, mildenhall2019local} have been proposed to address the problem and have achieved remarkable improvements. Recently, neural radiance fields \cite{martin2020nerf} demonstrates impressive results of view synthesis by representing scenes as continuous implicit radiance fields. It is further extended to operate on dynamic scenes \cite{park2020deformable, xian2020space}. \cite{liu2020neural} employs a sparse voxel octree and achieves great improvement over \cite{mildenhall2020nerf}. \cite{riegler2020free} employs image-based encoder-decoder architecture to process the proxy generated from the conventional sparse reconstruction, and is later improved by \cite{riegler2020stable}. While view synthesis is not the major focus of this work, we show that our guided optimization scheme consistently improves the synthesis quality of NeRF \cite{mildenhall2020nerf} on both seen and novel views, which shows the potential of using conventional sparse reconstructions to help improve the synthesis quality of NeRF-like methods.

%% file: figures/pipeline.tex
\begin{figure*}[tb]
	\centering
	\includegraphics[width=0.70\linewidth]{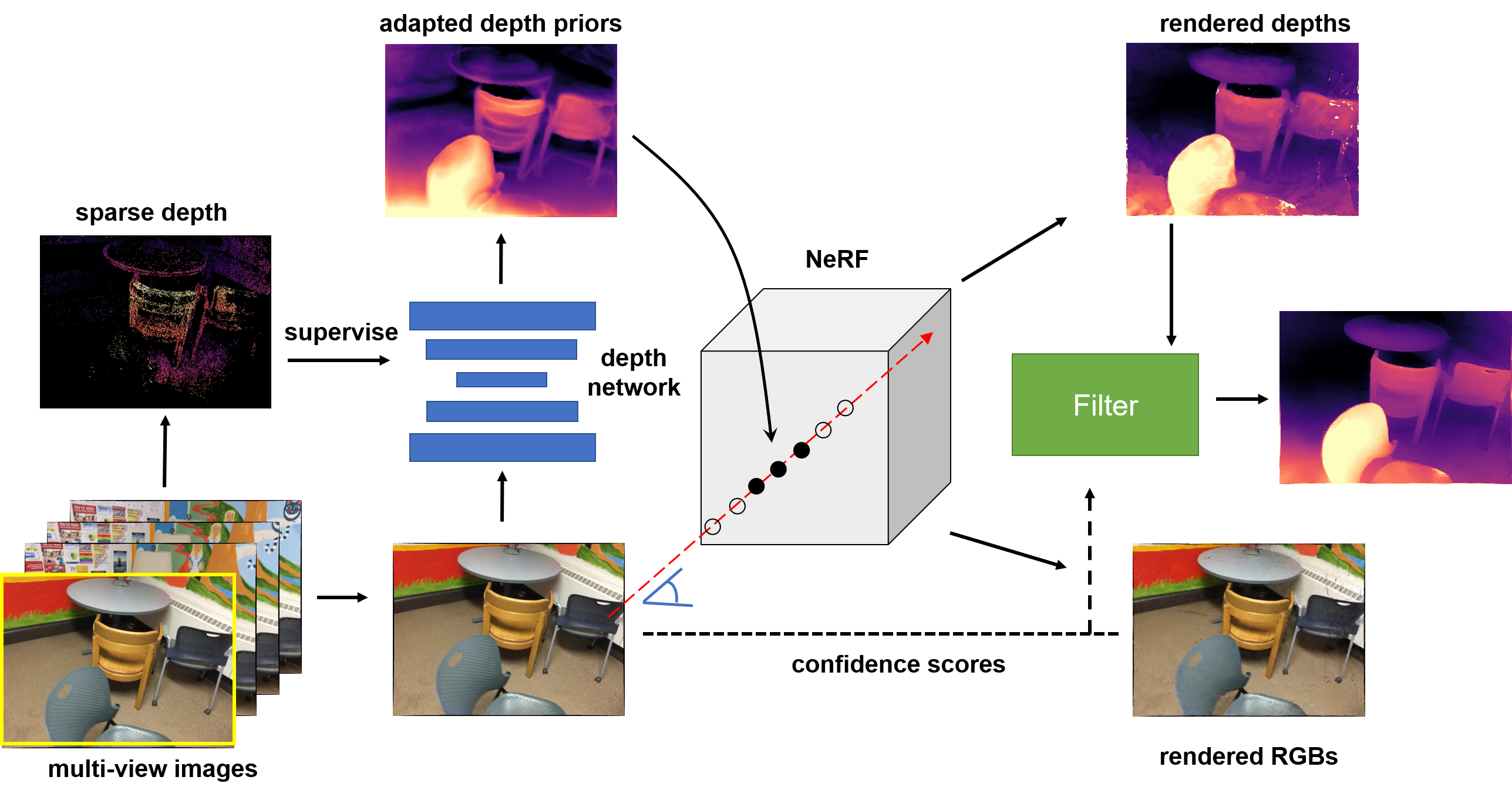}
	\caption{An overview of our method. We first adopt conventional SfM and MVS from COLMAP to get sparse depth (after fusion), which is used to train a monocular depth network to get scene-specific depth priors. Then, we utilize the depth priors to guide volume sampling in the optimization of NeRF \cite{mildenhall2020nerf}. Finally, by computing the errors between the rendered images and the original input images we acquire confidence scores, which enables us to employ a confidence-based filter to improve the rendered depths. }
	\label{fig:pipeline}
	\vspace{-15pt}
\end{figure*}

%% file: latex/method.tex
\subsection{Overview}
We introduce a multi-view depth estimation method that utilizes conventional sparse reconstruction and learning-based priors. Our proposed system builds on top of the recently proposed neural radiance fields (NeRF) \cite{mildenhall2020nerf} and performs test-time optimization at inference. Compared to the existing test-time optimization method \cite{luo2020consistent} that relies on estimated correspondences, directly optimizing over volumes eliminates the necessity of accurately matching cross-view pixels. This idea is also exploited by direct methods in the context of simultaneous localization and mapping (SLAM) \cite{engel2014lsd}.

The key to our approach is to effectively integrate the additional information from the learning-based priors into the NeRF training pipeline. Figure \ref{fig:pipeline} shows an overview of our proposed system. 
Section \ref{sec:3.2} shows how we adapt the depth priors to specific scenes at test time. In Section \ref{sec:3.3}, we analyze the reason why NeRF fails on producing accurate geometry in indoor scenes and describe our learning-based priors guided optimization scheme. In Section \ref{sec:3.4}, we discuss how to infer depth and synthesize views from the neural radiance fields trained with guided optimization.

\subsection{Scene-specific Adaptation of the Depth Priors}
\label{sec:3.2}
Similar to CVD \cite{luo2020consistent}, our method also aims to utilize learning-based depth priors to help optimize the geometry at test time. However, unlike \cite{luo2020consistent} that employs the same monocular depth network for all test scenes, we propose to adapt the network onto each scene to get scene-specific depth priors. Empirically this test-time adaptation method largely improves the quality of the final depth output. 

Our proposal on adapting scene-specific depth priors is to finetune a monocular depth network over its conventional sparse reconstruction. Specifically, we run COLMAP \cite{schonberger2016structure,schonberger2016pixelwise} on the test scene and acquire per-view sparse depth maps by projecting the fused 3D point clouds after multi-view stereo. Since geometric consistency check is adopted in the fusion step, the acquired depth map is sparse but robust and can be used as a supervision source for training the scene-specific depth priors.

Due to the scale ambiguity of acquired depth map, we employ the scale-invariant loss \cite{eigen2014depth} to train the depth network, which is written as follows:
\begin{equation}
\begin{aligned}
L(D^i_p, D^i_{Sparse})  = &  \frac{1}{n} \sum_{j=1}^n |\log D^i_p(j) - \log D^i_{Sparse}(j) \\ 
& + \alpha(D^i_p, D^i_{Sparse})|,
\end{aligned}
\end{equation}
where $D^i_p$ is the predicted depth map and $D^i_{Sparse}$ is the sparse depths acquired from COLMAP \cite{schonberger2016structure,schonberger2016pixelwise}. We align the scale of the predicted depth map with the sparse depth supervision by employing the scale factor $\alpha(D^i_p, D^i_{Sparse})$ in the loss formulation, which can be computed by averaging the difference over all valid pixels:
\begin{equation}
    \alpha(D^i_p, D^i_{Sparse}) = \frac{1}{n} \sum_j(\log D^i_p(j) - \log D^i_{Sparse}(j)).
\end{equation}

The finetuned monocular depth network is a stronger prior that fits the specific target scene. The quality of the adapted priors can be further improved with our guided optimization over NeRF, while Table \ref{tab:ablation_finetune} shows that applying existing correspondence-based neural optimization will surprisingly degrade the quality of the adapted depth priors.

\subsection{Guided Optimization of NeRF}
\label{sec:3.3}
Neural radiance fields were initially proposed in \cite{mildenhall2020nerf}, which achieves impressive results on view synthesis. Our system exploits its potential  for accurate depth estimation. By integrating the aforementioned adapted depth priors, we directly optimizate on implicit volumes. 
\input{figures/nerf-analysis}
The key to the success of NeRF is to employ a fully connected network parameterized by $\theta$ to represent implicit radiance fields with $F_{\theta} : (\mathbf{x}, \mathbf{d}) \to (\mathbf{c},\sigma)$, where $\mathbf{x}$ and $\mathbf{d}$ denotes the location and direction, $\mathbf{c}$ and $\sigma$ denotes the color and density as the network outputs. View synthesis can be easily achieved over NeRF with volume rendering, which enables NeRF to train itself directly over multi-view RGB images. During volume rendering, NeRF adopts the near bound $\timenear$ and the far bound $\timefar$ computed from the sparse 3D reconstruction to monitor the sampling space along each ray. Specifically, it partitions $[\timenear, \timefar]$ into $M$  bins and one query point is randomly sampled for each bin with a uniform distribution:
\begin{equation}
    t_i \sim \mathcal{U} \left[ \timenear + \frac{i-1}{M}(\timefar-\timenear),\,\, \timenear + \frac{i}{M}(\timefar-\timenear) \right].
    \label{eq:stratified}
\end{equation} 
The rendered RGB value $C(\mathbf{r})$ for each ray can be calculated from the finite samples with volume rendering. Moreover, per-view depth $D(\mathbf{r})$ can also be approximated by calculating the expectation of the samples along the ray:
\begin{equation}
\begin{aligned}
\label{eqn:render_coarse}
C(\mathbf{r})&=\sum_{i=1}^{M}T_i (1-\expo{-\absrp_i \deltatime_i}) c_i \\
D(\mathbf{r})&=\sum_{i=1}^{M}T_i (1-\expo{-\absrp_i \deltatime_i}) t_i
\end{aligned}
\end{equation}
where $T_i=\expo{- \sum_{j=1}^{i-1} \absrp_j \deltatime_j}$ indicates the accumulated transmittance from $\timenear$ to $t_i$ and $\deltatime_i = t_{i+1} - t_i $ is the distance between adjacent samples.

While simply satisfying the radiance field over the input image does not guarantee a correct geometry, the shape-radiance ambiguity between the 3D geometry and radiance has been studied in \cite{zhang2020nerf++}. It is believed in the paper that because incorrect geometry leads to high intrinsic complexity, the correct shape, with smoother surface light field, is more favored by the learned neural radiance fields with limited network capacity. This assumption generally holds for rich textured outdoor scenes. However, we empirically observe that NeRF struggles on poorly textured areas (e.g. walls), which are common in indoor environments. Figure \ref{fig::analysis} shows one failure case of NeRF that suffers from shape-radiance ambiguity in texture-less areas, where NeRF perfectly synthesizes the input image with a geometry largely deviated from the groundtruth. The failure comes from the fact that while extremely implausible shapes are ignored with the favor of smoothed surface light field \cite{zhang2020nerf++}, there still exists a family of smoothed radiance fields that perfectly explains the training images. Further, the blurred images and large-motion real-world indoor scenes will reduce the capacity of NeRF and aggravate the shape-radiance ambiguity issue. We find that this is a common issue in all tested indoor scenes. 

In Figure \ref{fig::analysis}(b), we show that all the sampled points along the camera ray that corresponds to a poorly textured pixel predict roughly the same RGB values, with the confidence distribution concentrated only in a limited range. Motivated by this observation, we consider guiding the NeRF sampling process with our adapted depth priors from the monocular depth network. By explicitly limiting the sampling range to be distributed around the depth priors, we avoid most degenerate cases for NeRF in indoor scenes. This enables accurate depth estimation by directly optimizing over RGB images. 

Specifically, we first acquire error maps of the adapted depth priors with a geometric consistency check. Denote the adapted depth priors as $\{D^i\}_{i=1}^{N}$ for the N input views. We project the depth map of each view to all the other views:
\begin{equation}
\label{eq:proj}
\begin{aligned}
p^{i \rightarrow j}, D^{i \rightarrow j} &= proj(K, T^{i\rightarrow j}, D^i) \\
D^{j'} &= D^j(p^{i \rightarrow j}),
\end{aligned}
\end{equation}
where $K$ is the camera intrinsics, $T^{i\rightarrow j}$ is the relative pose. $p^{s \rightarrow t}$ and $D^{i \rightarrow j}$ are the 2D coordinates and depth of the projection in $j$th view. Then we calculate the depth reprojection error using the relative error between $D^{j'}$ and $D^{i \rightarrow j}$. Note that there are pixels that do not overlap across some view pairs. Thus, we define the error map of the depth priors for each view $e_i$ as the average value of the top $K$ minimum cross-view depth projection error.
\input{figures/depth_prior}

We use the error maps $\{e^i\}_{i=1}^{N}$ to calculate adaptive sample ranges $[t_n, t_f]$ for each camera ray:
\begin{equation}
\begin{aligned}
\label{eq:range}
\timenear &= D(1- clamp(e,\alpha_{l},\alpha_{h})) \\
\timefar &= D(1+ clamp(e,\alpha_{l},\alpha_{h}))
\end{aligned}
\end{equation}
where $\alpha_l$ and $\alpha_h$ defines the relative lower and higher bounds of the ranges. With the adaptive ranges we achieve a balance between diversity and precision of the confidence distribution along camera rays. As illustrated in Figure \ref{fig:prior}, the sampling over pixels with relatively low error is more concentrated around the adapted depth priors, while the sampling over pixels with large error is close to the original NeRF formulation.

\subsection{Inference and View Synthesis}
\label{sec:3.4}
For inference, we can directly predict the depth map for each input view by resampling within the sampling range defined in Eq. (\ref{eq:range}) and applying Eq. (\ref{eq:stratified}) to compute the expectation. This gives an accurate output depth for the NeRF equipped with our proposed guided optimization scheme. 

To further improve depth quality, we exploit the potential of using the view synthesis results of NeRF to compute per-pixel confidence for the predicted geometry. If the rendered RGB at a specific pixel does not match the input training image well, we attach a relatively low confidence for the depth prediction of this pixel. The confidence $S^i_j$ for the $j$th pixel in the $i$th view is specifically defined as:
\begin{equation}
\label{eq:confidence}
\begin{aligned}
S^i_j = 1 - \frac{1}{3}||C_{gt}^i(j) - C_{render}^i(j)||_1,
\end{aligned}
\end{equation}
where $C_{gt}^i$ and $C_{render}^i$ are the groundtruth images and rendered images for each seen view with all the values divided by 255. The absolute difference is employed.
\input{tables/main-depth}
This confidence map can be further used to refine the predicted depth map with off-the-shelf post-filtering techniques. We employ plane bilateral filtering introduced in \cite{valentin2018depth} over the depth to get the final output, which improves depth quality especially for the regions where rendered RGB images are not accurate. 

While the proposed guided optimization strategy needs the adapted depth priors as input to guide point sampling along the camera ray, we can still perform novel view synthesis by directly using the adapted depth priors from the nearest seen view. Empirically this is sufficient to produce accurate depth maps and significantly outperforms the original NeRF in terms of view synthesis quality (See Table \ref{tab:synthesis}).

%% file: figures/nerf-analysis.tex
\begin{figure}[tb]
\centering
\vspace{-5pt}
\setlength\tabcolsep{1.0pt} 
\renewcommand{\arraystretch}{1.0}
\begin{tabular}{cc}
{\includegraphics[width=0.45\linewidth]{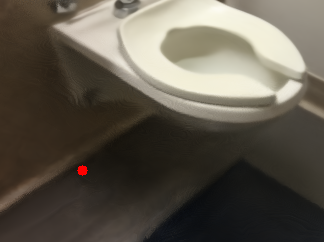}} & {\includegraphics[width=0.45\linewidth]{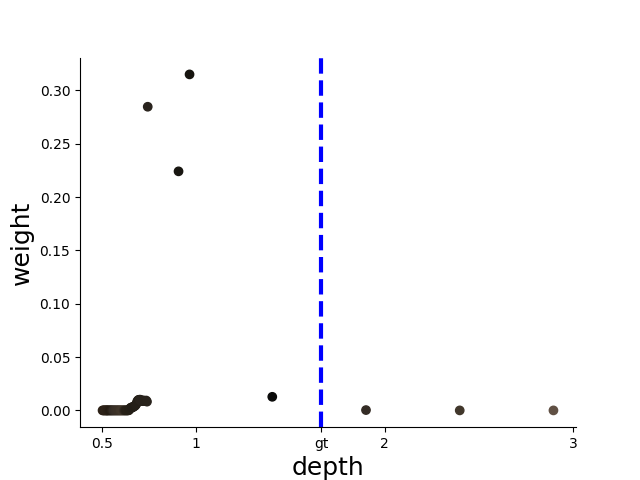}}

 \\
 (a) rendered RGB & (b) sampled points\\
{\includegraphics[width=0.45\linewidth]{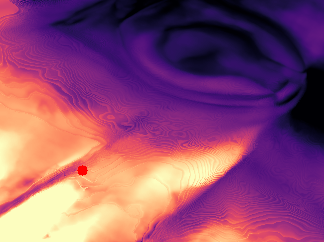}}   & 
{\includegraphics[width=0.45\linewidth]{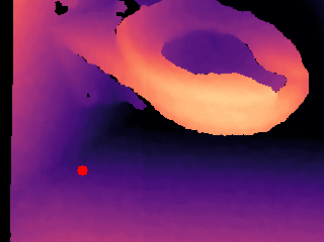}} \\
(c) rendered depth & (d) GT depth

\end{tabular}
\centering
\captionof{figure}{The inherent shape-radiance ambiguity \cite{zhang2020nerf++} becomes a bottleneck in indoor scenes. \textbf{Top row:} (a) rendered RGB of NeRF \cite{mildenhall2020nerf}. (b) visualization of the sampled points along the camera ray at the position colored in red. The blue line indicates the groundtruth depth value. \textbf{Bottom row:} (c) the rendered depth map of NeRF \cite{mildenhall2020nerf}. (d) the groundtruth depth map. While NeRF produces high quality rendered image (PSNR: 31.53), the rendered depth largely deviates from the groundtruth. 
}
\vspace{-10pt}
\label{fig::analysis}
\end{figure}

%% file: figures/depth_prior.tex
\begin{figure}[tb]
	\centering
	\includegraphics[width=1.0\linewidth]{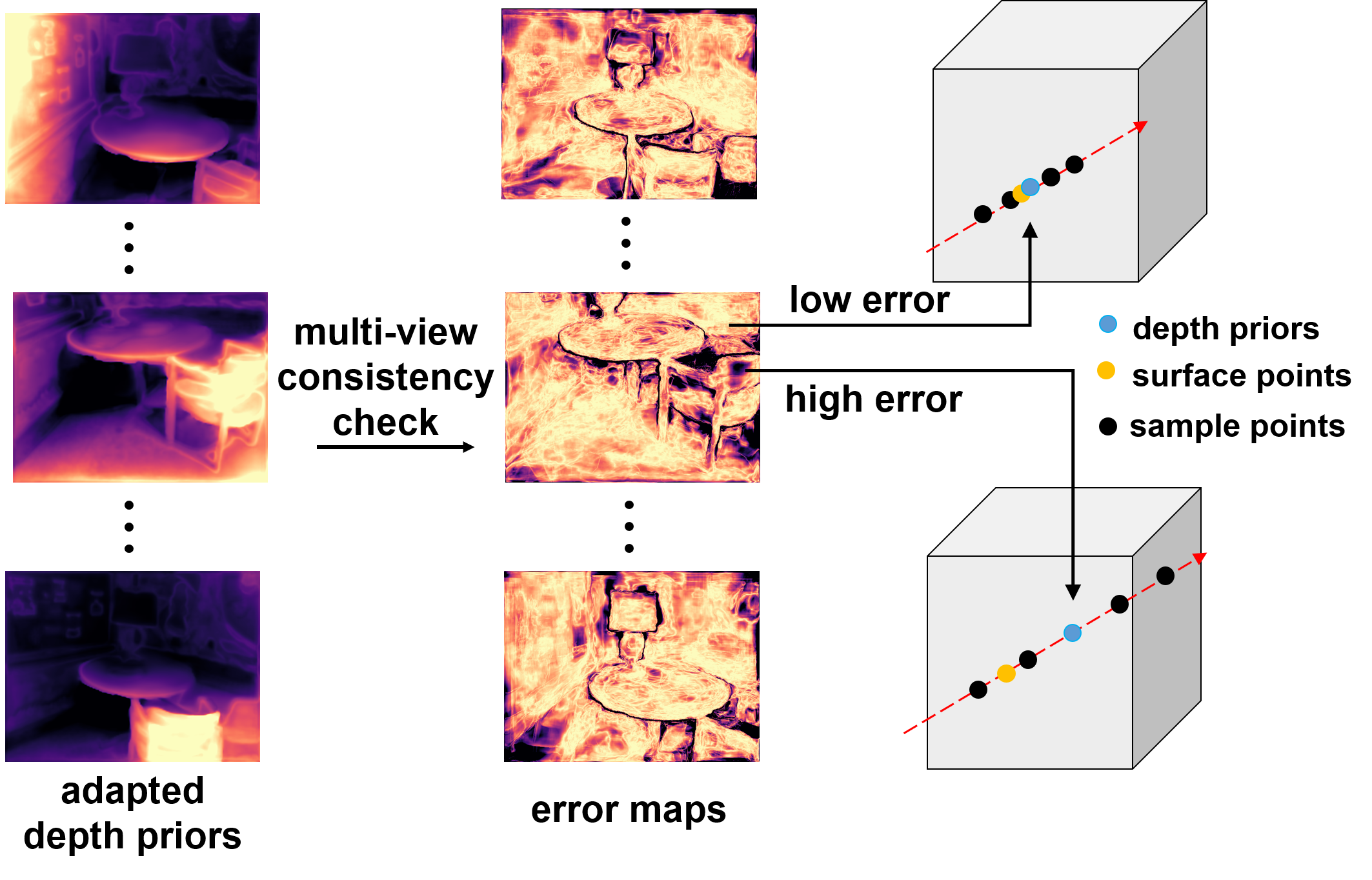}
	\caption{Guided optimization of NeRF \cite{mildenhall2020nerf}. We adopt multi-view consistency check on adapted depth priors to get error maps, which help calculate adaptive depth ranges for each camera ray to sample points for NeRF optimization.}
	\label{fig:prior}
	\vspace{-10pt}
\end{figure}

%% file: tables/main-depth.tex
\begin{table*}[tb]
	\centering
	\resizebox{0.9\textwidth}{!}{
		\begin{tabular}{|l||c|c|c|c|c|c|c|}
			\hline
			Method &\cellcolor{col1}Abs Rel & \cellcolor{col1}Sq Rel & \cellcolor{col1}RMSE  & \cellcolor{col1}RMSE log & \cellcolor{col2}$\delta < 1.25 $ & \cellcolor{col2}$\delta < 1.25^{2}$ & \cellcolor{col2}$\delta < 1.25^{3}$\\
			\hline
			COLMAP \cite{schonberger2016structure,schonberger2016pixelwise}& 0.4619 &0.6308 &1.0125 &1.7345 &0.4811 &0.5139 &0.5333 
\\

			ACMP \cite{xu2020planar}&  0.1945 &0.1710 &0.4551 &0.3056 &0.7309 &0.8810 &0.9419 

 \\			\hline
			DELTAS \cite{sinha2020deltas}&0.1001 &0.0319 &0.2070 &0.1284 &0.8618 &0.9920 &0.9991 
  \\
			Atlas \cite{murez2020atlas}&0.0776 &0.0631 &0.2441 &0.2693 &0.9289 &0.9536 &0.9594 
 \\
			DeepV2D \cite{teed2018deepv2d} & 0.0818 &0.0226 &0.1714 &0.1095 &0.9414 &0.9908 &0.9979 
  \\			\hline
  			NeRF \cite{mildenhall2020nerf}&0.3929 &1.4849 &1.0901 &0.5210 &0.4886 &0.7318 &0.8285 

\\
			Mannequin \cite{li2019learning}&0.1554 &0.0636 &0.2969 &0.1806 &0.7859 &0.9735 &0.9953 

 \\
			CVD \cite{luo2020consistent}&0.0995 &0.0304 &0.1945 &0.1269 &0.9008 &0.9879 & 0.9971 \\
            Ours w/o filter & 0.0635 &0.0145 &0.1455 &0.0936 &0.9541 &0.9910 &0.9989 \\ 
			Ours & \bf{0.0614} & \bf{0.0126} &\bf{0.1345} &\bf{0.0861}&\bf{0.9601} &\bf{0.9955} &\bf{0.9996} 
  \\
			\hline
	\end{tabular}}
	\caption{Quantitative comparisons for multi-view depth estimation.  Scores are averaged over 8 scenes from ScanNet. 
	}
	\vspace{-2mm}
	\label{tab:depth}
\end{table*}

%% file: latex/exp.tex
\subsection{Experimental Setup}
\noindent \textbf{Dataset:}
We conducted experiments on ScanNet \cite{dai2017scannet} dataset.  Following the experimental setup in NeRF \cite{mildenhall2020nerf}, we randomly selected 8 scenes to evaluate our method. For each scene, we picked 40 images covering a local region and held out $1/8$ of these as the test set for novel view synthesis. All images are resized as $484 \times 648$ resolution. Due to the scale ambiguity issue, we adopted the median groundtruth scaling strategy \cite{zhou2017unsupervised} for depth evaluation. 

\noindent \textbf{Implementation Details:}
For the adapted depth priors, following CVD \cite{luo2020consistent}, we used the network architecture introduced in Mannequin Challenge \cite{li2019learning} with its pretrained weights as our monocular depth network. 15 finetuning epochs were used in the scene-specific adaptation. We set $K=4$ for multi-view consistency check and $\alpha_{l}=0.05,\alpha_{h}=0.15$ as the bounds of sample ranges. Please refer to our supplementary material for more details. 
\input{tables/ft-depth}
\input{figures/ft-depth}

\subsection{Results on Multi-view Depth Estimation}
Table \ref{tab:depth} shows the results for depth estimation task on ScanNet \cite{dai2017scannet}. For all methods, we used their released implementation in the experiments. We also report results without applying the filtering step.
\input{figures/main-depth}
Our method outperforms state-of-the-art depth estimation methods in all metrics. Note that DeepV2D \cite{teed2018deepv2d}, DELTAS \cite{sinha2020deltas} and Atlas \cite{murez2020atlas} are all trained on ScanNet with groundtruth depth supervision. With the proposed guided optimization scheme, our method mitigates the problem of the shape-radiance ambiguity and demonstrates the potential of exploiting NeRF for accurate depth estimation. Figure \ref{fig::depth_main} shows some qualitative results. While the original NeRF \cite{mildenhall2020nerf} fails to predict reasonable geometry, our method generates visually appealing depth maps. The confidence-based filter can further refine the predicted depth by smoothing the per-pixel estimation of NeRF \cite{mildenhall2020nerf}.

\input{tables/ablation-full}
To further study the advantages of optimizing over implicit volumes, we also applied the optimization of CVD \cite{luo2020consistent} on our adapted depth priors. Results are shown in Table \ref{tab:ablation_finetune} and one example is exhibited in Figure \ref{fig::depth_ft}. We surprisingly find that the optimization of CVD degrades the depth quality of the initial depth priors. This is mainly due to wrong estimated correspondences from the employed flow network in \cite{luo2020consistent}. The flow estimation is particularly challenging over poorly textured regions, which is ubiquitous in indoor scenes. The proposed guided optimization enables us to integrate depth priors on top of NeRF \cite{mildenhall2020nerf}, which directly optimizes on raw RGB images, avoiding the challenging step of correspondence estimation in indoor scenes.

\subsection{Ablation Studies}
To better understand the working mechanism of our method, we performed ablation studies over each component of the proposed system. Results in Table \ref{tab:ablation_full} show that each component is beneficial to the final depth quality. This verifies the advantages of integrating depth priors into the optimization of NeRF \cite{mildenhall2020nerf}. 

\input{tables/ablation-guided}

\input{tables/main-synthesis}

We further study the design of adaptive ranges used in the guided optimization. It is shown that both the adaptive strategy and the use of bounds contribute to the performance gain. With the computed error maps, $\alpha_l$ and $\alpha_h$ avoid the samples being over-concentrated or overly random respectively, which enables the sampling to reach a balance between diversity and precision of the sampled points.

\subsection{Results on View Synthesis}
We also observe that the proposed guided optimization scheme is beneficial to the view synthesis quality of NeRF. Figure \ref{fig::synthesis} illustrates some visualizations. Table \ref{tab:synthesis} shows results on novel view synthesis, where our method consistently improves NeRF on all 8 scenes. Although view synthesis is not the main focus of our work, we achieve comparable or even better results compared to state-of-the-art novel view synthesis methods \cite{riegler2020stable,bi2020neural}. Note that SVS \cite{riegler2020stable} employs image-based novel view synthesis methods over the information extracted from sparse reconstruction. Our method, with the guided optimization scheme, opens a new way to employ the robust conventional sparse reconstruction to improve the synthesis quality directly over implicit 3D volumes. In addition, results in Table \ref{tab:seen} show that our method can improve the view synthesis quality of NeRF on seen views. The guided optimization helps NeRF to focus on more informative regions and improves its capacity for rendering RGB images.

\input{tables/seen-view}
\input{figures/synthesis}

%% file: tables/ft-depth.tex
\begin{table}[tb]
	\centering
	\resizebox{0.48\textwidth}{!}{
		\begin{tabular}{|c||c|c|c|} 
			\hline
			\tabincell{c}Method& \cellcolor{col1}Abs Rel  & \cellcolor{col1}Sq Rel & \cellcolor{col2}$\delta < 1.25 $\\
			\hline
			Mannequin \cite{li2019learning} &0.1554 &0.0636 & 0.7859 \\
			adapted depth priors & 0.0844&  0.0223&   0.9410\\
			CVD optimization  & 0.0886&  0.0251&   0.9128\\
			Our optimization  & \bf{0.0635}&  \bf{0.0145}&   \bf{0.9541}\\
			
			\hline 
	\end{tabular}}
	\caption{Comparison of the effectiveness of test-time optimization between CVD \cite{luo2020consistent} and our method. Both methods perform optimization over the adapted depth priors, which is acquired by training the Mannequin Challenge depth network \cite{li2019learning} with sparse supervision from COLMAP \cite{schonberger2016structure,schonberger2016pixelwise}. Scores are averaged over 8 scenes.}
	\vspace{-3mm}
	\label{tab:ablation_finetune}
\end{table}

%% file: figures/ft-depth.tex
\begin{figure}[tb]
\centering
\vspace{-5pt}
\setlength\tabcolsep{1.0pt} 
\renewcommand{\arraystretch}{1.0}
\begin{tabular}{cccc}
{\includegraphics[width=0.25\linewidth]{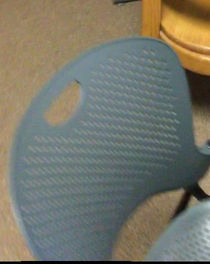}} & 
{\includegraphics[width=0.25\linewidth]{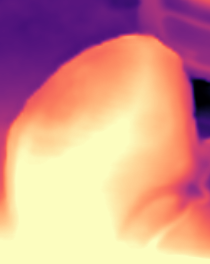}} & 
{\includegraphics[width=0.25\linewidth]{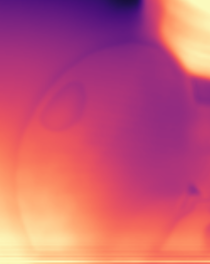}} &
{\includegraphics[width=0.25\linewidth]{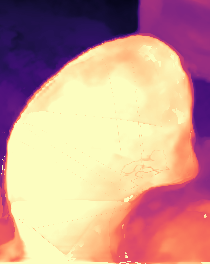}} \\
RGB & \tabincell{c}{adapted \\ depth priors} & \tabincell{c}{CVD \\ optimization} & \tabincell{c}{Our \\ optimization} \\
\end{tabular}
\centering
\captionof{figure}{The optimization of CVD \cite{luo2020consistent} surprisingly degrades the quality of the depth priors due to unreliable flow correspondences, while our method achieves improvement with guided optimization of NeRF \cite{mildenhall2020nerf} }
\label{fig::depth_ft}
\vspace{-5mm}
\end{figure}

%% file: figures/main-depth.tex
\begin{figure*}[tb]
\centering
\vspace{-5pt}
\setlength\tabcolsep{1.0pt} 
\renewcommand{\arraystretch}{1.0}
\begin{tabular}{cccccccc}
{\includegraphics[width=0.12\linewidth]{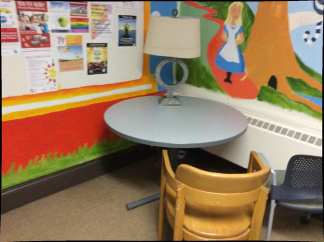}} & 
{\includegraphics[width=0.12\linewidth]{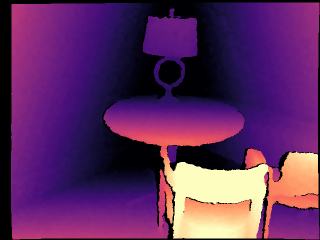}} & 
{\includegraphics[width=0.12\linewidth]{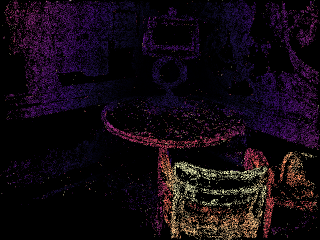}} & 
{\includegraphics[width=0.12\linewidth]{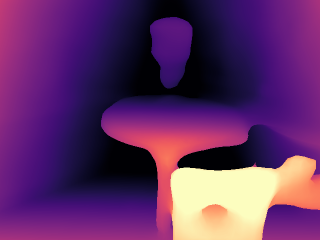}} &
{\includegraphics[width=0.12\linewidth]{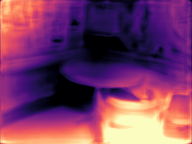}}&
{\includegraphics[width=0.12\linewidth]{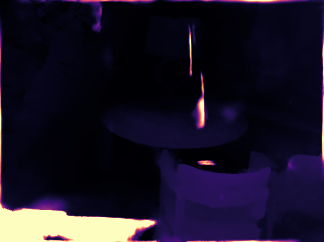}} & 
{\includegraphics[width=0.12\linewidth]{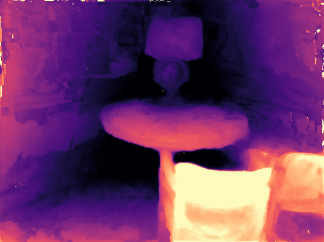}} &
{\includegraphics[width=0.12\linewidth]{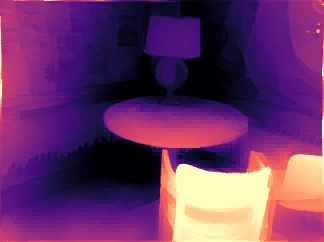}} \\
{\includegraphics[width=0.12\linewidth]{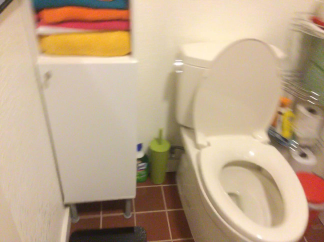}} & 
{\includegraphics[width=0.12\linewidth]{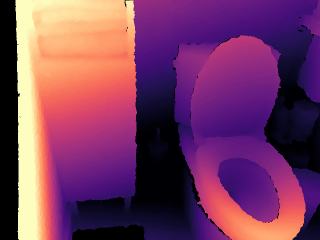}} & 
{\includegraphics[width=0.12\linewidth]{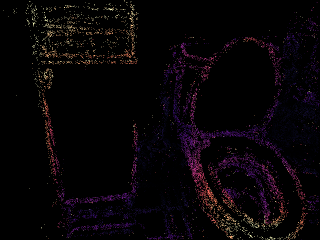}} & 
{\includegraphics[width=0.12\linewidth]{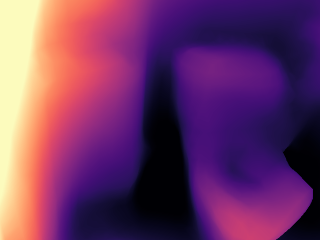}} &
{\includegraphics[width=0.12\linewidth]{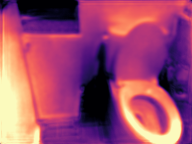}}&
{\includegraphics[width=0.12\linewidth]{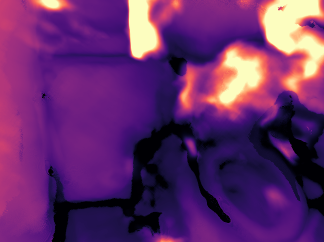}} & 
{\includegraphics[width=0.12\linewidth]{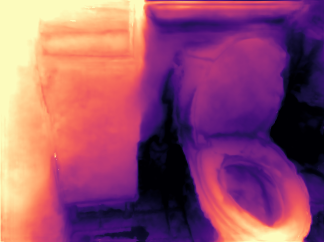}} &
{\includegraphics[width=0.12\linewidth]{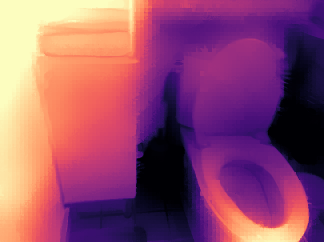}} \\
{\includegraphics[width=0.12\linewidth]{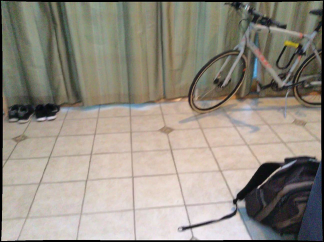}} & 
{\includegraphics[width=0.12\linewidth]{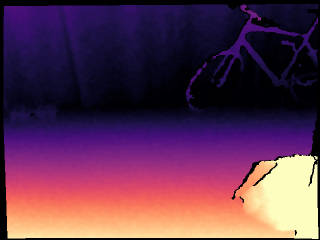}} & 
{\includegraphics[width=0.12\linewidth]{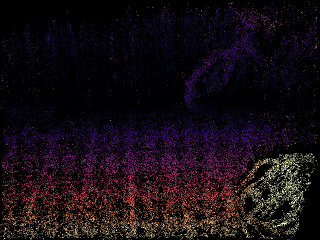}} & 
{\includegraphics[width=0.12\linewidth]{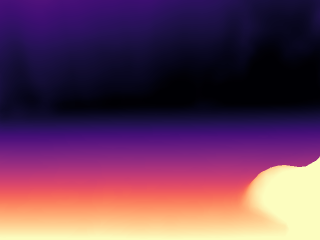}} &
{\includegraphics[width=0.12\linewidth]{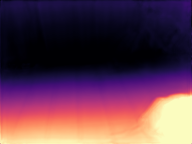}}&
{\includegraphics[width=0.12\linewidth]{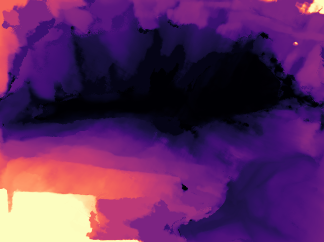}} & 
{\includegraphics[width=0.12\linewidth]{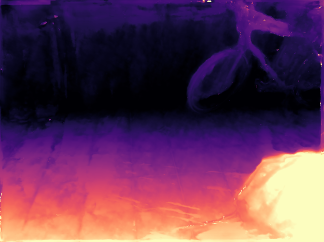}} &
{\includegraphics[width=0.12\linewidth]{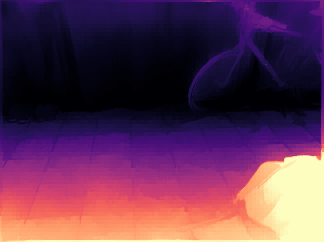}} \\
{\includegraphics[width=0.12\linewidth]{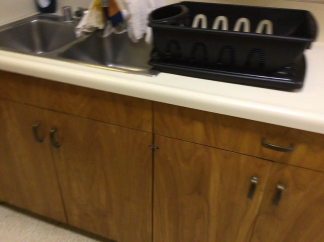}} & 
{\includegraphics[width=0.12\linewidth]{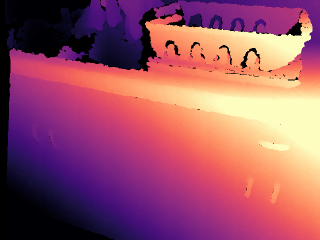}} & 
{\includegraphics[width=0.12\linewidth]{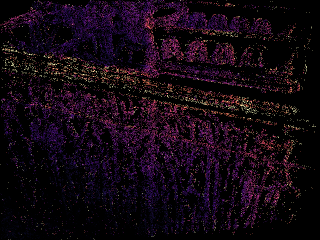}} & 
{\includegraphics[width=0.12\linewidth]{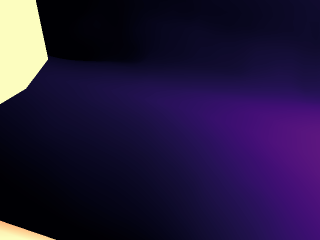}} &
{\includegraphics[width=0.12\linewidth]{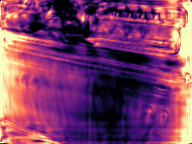}}&
{\includegraphics[width=0.12\linewidth]{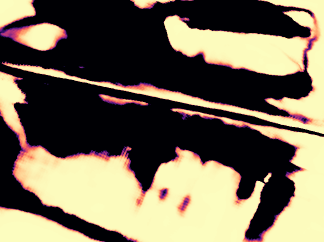}} & 
{\includegraphics[width=0.12\linewidth]{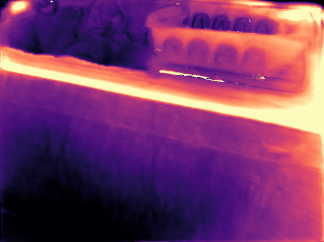}} &
{\includegraphics[width=0.12\linewidth]{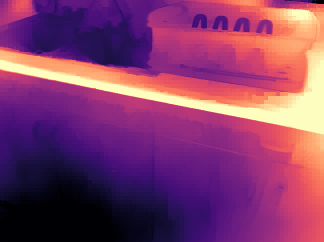}} \\
{\includegraphics[width=0.12\linewidth]{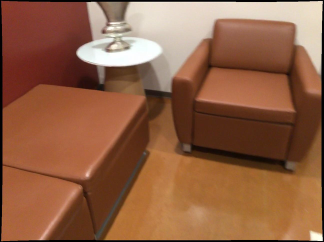}} & 
{\includegraphics[width=0.12\linewidth]{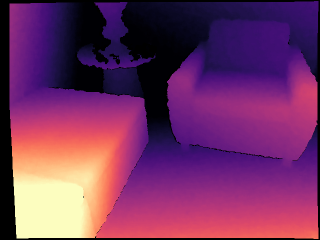}} & 
{\includegraphics[width=0.12\linewidth]{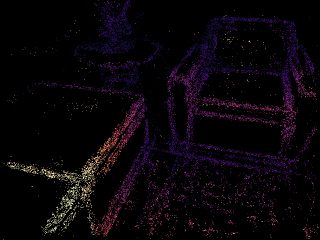}} & 
{\includegraphics[width=0.12\linewidth]{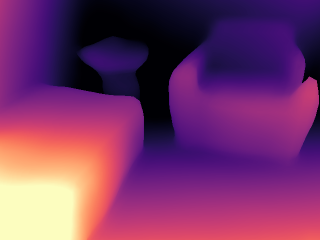}} &
{\includegraphics[width=0.12\linewidth]{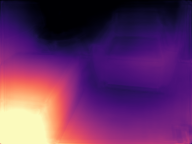}}&
{\includegraphics[width=0.12\linewidth]{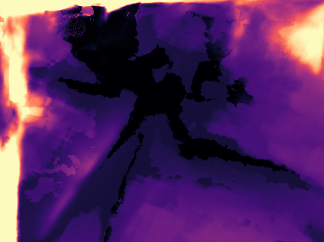}} & 
{\includegraphics[width=0.12\linewidth]{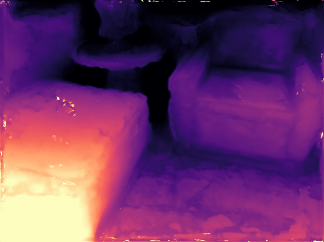}} &
{\includegraphics[width=0.12\linewidth]{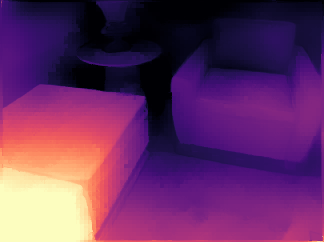}} \\
{\includegraphics[width=0.12\linewidth]{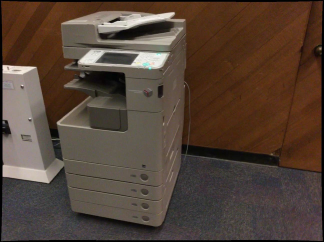}} & 
{\includegraphics[width=0.12\linewidth]{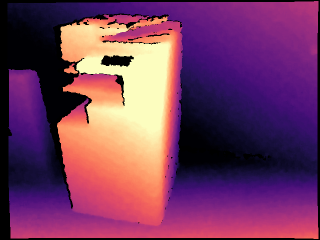}} & 
{\includegraphics[width=0.12\linewidth]{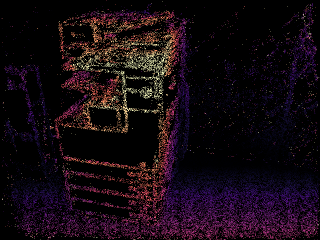}} & 
{\includegraphics[width=0.12\linewidth]{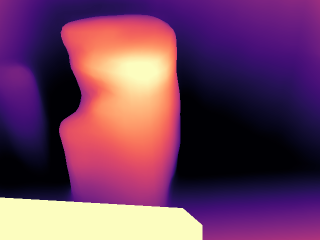}} &
{\includegraphics[width=0.12\linewidth]{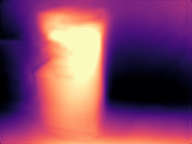}}&
{\includegraphics[width=0.12\linewidth]{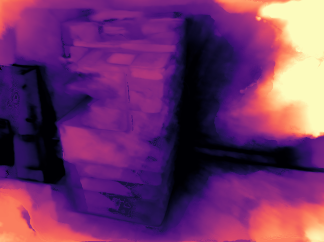}} & 
{\includegraphics[width=0.12\linewidth]{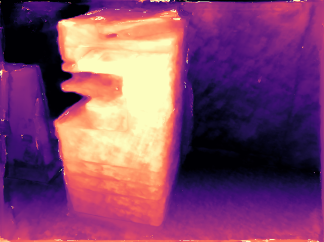}} &
{\includegraphics[width=0.12\linewidth]{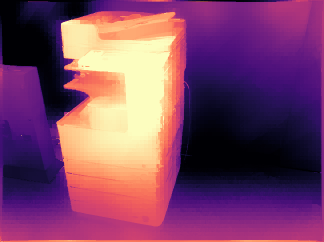}} \\
RGB & GT depths& COLMAP \cite{schonberger2016pixelwise} & Atlas \cite{murez2020atlas}& CVD \cite{luo2020consistent} & NeRF \cite{mildenhall2020nerf} & Ours w/o filter &Ours \\
\end{tabular}
\centering
\captionof{figure}{Qualitative comparisons on ScanNet \cite{dai2017scannet} dataset. Our method, without the post-filtering step, outperforms all compared methods in terms of depth quality. The filter further smooths the per-pixel estimated depth maps. \textbf{Better viewed when zoomed in.}}
\label{fig::depth_main}
\vspace{-5mm}
\end{figure*}

%% file: tables/ablation-full.tex
\begin{table}[tb]
	\centering
	\resizebox{0.47\textwidth}{!}{
		\begin{tabular}{|c|c|c||c|c|c|} 
			\hline
			NeRF & \tabincell{c}{depth \\ priors} & filter& \cellcolor{col1}Abs Rel  & \cellcolor{col1}Sq Rel & \cellcolor{col2}$\delta < 1.25 $\\
			\hline
			\checkmark&  &  & 0.302&  0.210&   0.518\\
			& \checkmark &  & 0.067&  0.010&   0.960\\
			\checkmark&  &  \checkmark& 0.287&  0.167&   0.546\\
			& \checkmark &  \checkmark& 0.065&  0.009&   0.966\\
			\checkmark& \checkmark &  & 0.053&  0.006&   0.979\\
			\checkmark& \checkmark &  \checkmark& \bf{0.051}&  \bf{0.005}&   \bf{0.987} \\
			\hline 
	\end{tabular}}
	\caption{Ablation studies on each component of our system. For the experiments `NeRF + filter' and `depth priors + filter', we compute the confidence scores by using the relative errors between the prediction depths and the groundtruth. The experiment was conducted on scene0521.}
	\vspace{-3mm}
	\label{tab:ablation_full}
\end{table}

%% file: tables/ablation-guided.tex
\begin{table}[tb]
	\centering
	\resizebox{0.47\textwidth}{!}{
		\begin{tabular}{|c|c||c|c|c|} 
			\hline
			\tabincell{c}adaptive & bound& \cellcolor{col1}Abs Rel  & \cellcolor{col1}Sq Rel & \cellcolor{col2}$\delta < 1.25 $\\
			\hline
	         &  & 0.065&  0.009&   0.971\\
			 \checkmark &  & 0.056&  0.009&   0.978\\
			\checkmark& \checkmark   & \bf{0.051}&  \bf{0.005}&   \bf{0.987}\\
			\hline 
	\end{tabular}}
	\caption{Ablation studies on the design of the proposed guided optimization with adaptive ranges. `bound' denotes the use of $\alpha_{l}$ and $\alpha_{h}$ in Eq. (\ref{eq:range}). For the experiment without using adaptive depth ranges for each camera ray, we set a fixed relative depth range to $[0.9, 1.1]$. The experiment was conducted on scene0521.}
	\vspace{-3mm}
	\label{tab:ablation_guided}
\end{table}

%% file: tables/main-synthesis.tex
\begin{table*}[t!]
	\centering
	\resizebox{0.85\textwidth}{!}{
		\begin{tabular}{|l||cc|cc|cc|cc|}
		\hline
			\multirow{2}{*}{Method}  & \multicolumn{2}{c|}{scene 0616} &\multicolumn{2}{c|}{scene 0521} &\multicolumn{2}{c|}{scene 0000} &\multicolumn{2}{c|}{scene 0158}  \\   
			 &  \multicolumn{1}{c}{PSNR$\uparrow$} & \multicolumn{1}{c|}{SSIM$\uparrow$} &  \multicolumn{1}{c}{PSNR$\uparrow$} & \multicolumn{1}{c|}{SSIM$\uparrow$} & \multicolumn{1}{c}{PSNR$\uparrow$} & \multicolumn{1}{c|}{SSIM$\uparrow$} & \multicolumn{1}{c}{PSNR$\uparrow$} & \multicolumn{1}{c|}{SSIM$\uparrow$}    \\   \hline
			NSVF \cite{liu2020neural}& 15.71&0.704    &27.73    &0.892     &\bf{23.36}&0.823     &\bf{31.98}&\bf{0.951} \\
			SVS \cite{riegler2020stable}& \bf{21.38}&\bf{0.899}&\bf{27.97}&\bf{0.924} &21.39     &\bf{0.914}&29.43&\bf{0.953} \\ \hline
			NeRF \cite{mildenhall2020nerf}& 15.76    & 0.699    &	24.41     &0.871     & 18.75    & 0.751    &29.19&0.928  \\
			Ours& 18.07    &0.748     &\bf{28.07} &0.901     &22.10     &0.880     &30.55&\bf{0.948} \\

			\hline
			
			\multirow{2}{*}{Method}  & \multicolumn{2}{c|}{scene 0316} &\multicolumn{2}{c|}{scene 0553} &\multicolumn{2}{c|}{scene 0653} &\multicolumn{2}{c|}{scene 0079}  \\   
			 &  \multicolumn{1}{c}{PSNR$\uparrow$} & \multicolumn{1}{c|}{SSIM$\uparrow$} &  \multicolumn{1}{c}{PSNR$\uparrow$} & \multicolumn{1}{c|}{SSIM$\uparrow$} & \multicolumn{1}{c}{PSNR$\uparrow$} & \multicolumn{1}{c|}{SSIM$\uparrow$} & \multicolumn{1}{c}{PSNR$\uparrow$} & \multicolumn{1}{c|}{SSIM$\uparrow$}    \\   \hline
			
			NSVF \cite{liu2020neural}& \bf{22.29}&0.917&31.15&0.947&	28.95&0.929&	26.88&0.887 \\
			SVS \cite{riegler2020stable}& 20.63&\bf{0.941}&30.95&\bf{0.968}&27.91&\bf{0.965}&25.18&\bf{0.923} \\ \hline
			NeRF \cite{mildenhall2020nerf}& 17.09	&0.828&30.76&0.950&	30.89&0.953	&25.48&0.896
 \\
			Ours& 20.88&0.899&	\bf{32.56}&\bf{0.965}&\bf{31.43}&\bf{0.964}&\bf{27.27}&\bf{0.916} \\
			\hline
	\end{tabular}}
	\caption{Quantitative comparisons for novel view synthesis. 
Numbers in bold are within 1\% of the best. }
	\vspace{-2mm}
	\label{tab:synthesis}
\end{table*}

%% file: tables/seen-view.tex
\begin{table}[t!]
	\centering
	\resizebox{0.4\textwidth}{!}{
		\begin{tabular}{|l||ccc|}
		\hline
		Method & PSNR $\uparrow$ & SSIM $\uparrow$ & LPIPS $\downarrow$\\
		\hline
		NeRF \cite{mildenhall2020nerf}&28.62&0.909&0.319 \\ 
		Ours&\bf{31.55}& \bf{0.942}& \bf{0.200} \\
		\hline
	\end{tabular}}
	\caption{Comparison between NeRF \cite{mildenhall2020nerf} and our method on seen views. Results are averaged over 8 scenes. 
}
	\vspace{-4mm}
	\label{tab:seen}
\end{table}

%% file: figures/synthesis.tex
\begin{figure}[tb]
\centering
\vspace{-5pt}
\setlength\tabcolsep{1.0pt} 
\renewcommand{\arraystretch}{1.0}
\begin{tabular}{ccc}
{\includegraphics[width=0.33\linewidth]{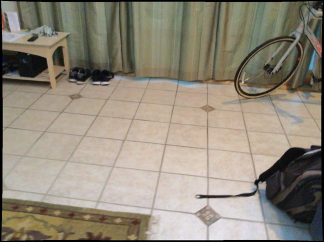}} & 
{\includegraphics[width=0.33\linewidth]{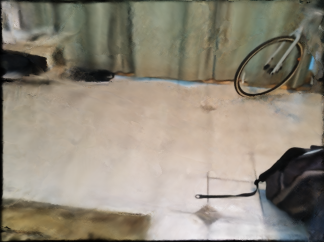}} & 
{\includegraphics[width=0.33\linewidth]{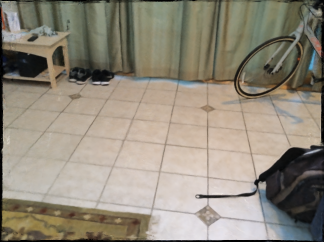}} \\
{\includegraphics[width=0.33\linewidth]{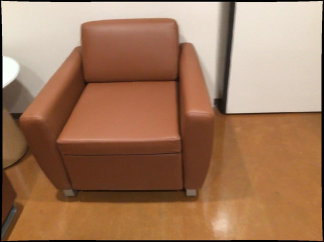}} & 
{\includegraphics[width=0.33\linewidth]{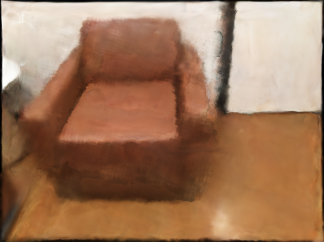}} & 
{\includegraphics[width=0.33\linewidth]{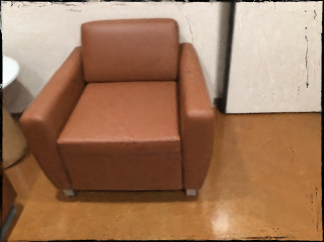}} \\
{\includegraphics[width=0.33\linewidth]{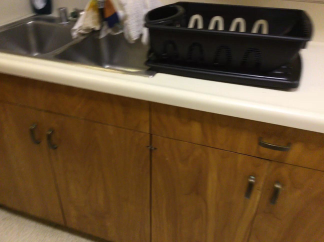}} & 
{\includegraphics[width=0.33\linewidth]{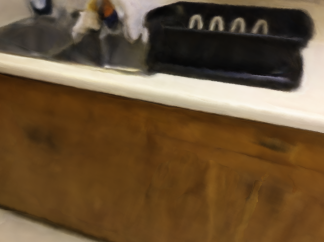}} & 
{\includegraphics[width=0.33\linewidth]{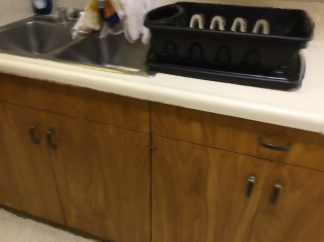}} \\
{\includegraphics[width=0.33\linewidth]{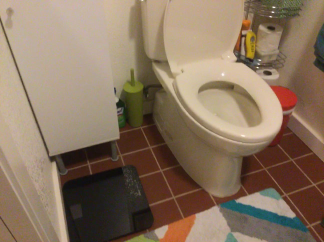}} & 
{\includegraphics[width=0.33\linewidth]{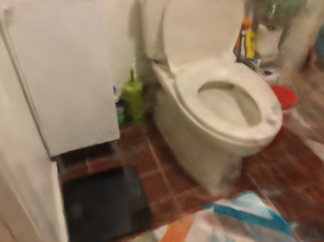}} & 
{\includegraphics[width=0.33\linewidth]{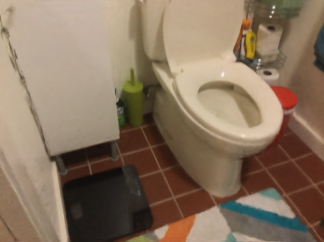}} 
\\
GT & NeRF \cite{mildenhall2020nerf} & Ours \\
\end{tabular}
\centering
\captionof{figure}{Results on view synthesis. The top two rows are rendering results on seen (training) views while the bottom two rows are on the novel views. With the adapted depth priors, our method improves the rendering quality for both seen and novel views. \textbf{Better viewed when zoomed in.}}
\label{fig::synthesis}
\vspace{-5pt}
\end{figure}

%% file: latex/conclusion.tex
In this work, we present a multi-view depth estimation method that integrates learning-based depth priors into the optimization of NeRF. Contrary to existing studies, we show that the shape-radiance ambiguity of NeRF becomes a bottleneck for NeRF-based depth estimation in indoor scenes. To address the issue, we propose a guided optimization framework to regularize the sampling process of NeRF during volume rendering with the adapted depth priors. Our proposed system demonstrates the significant improvement over prior works for indoor multi-view depth estimation, with a surprising finding that correspondence-based optimization can degrade the quality of depth priors in indoor scenes due to wrongly estimated flow correspondence. In addition, we also observe that the guided optimization improves the view synthesis quality of NeRF. 

While our optimization is 3x faster than NeRF due to the advantages of guided optimization, the current method is still not efficient and thus hard to be scaled up to large datasets. Nonetheless, our work demonstrates the potential of using neural radiance fields for accurate depth estimation. Future work includes efficient optimization, non-rigid reconstruction and visual effects based on the improved geometric structure in the learned neural radiance fields.

%% file: latex/acknowledgement.tex
This work was supported in part by the National Natural Science Foundation of China under Grant U1813218, Grant U1713214, and Grant 61822603, in part by a grant from the Beijing Academy of Artificial Intelligence (BAAI), and in part by a grant from the Institute for Guo Qiang, Tsinghua University.

%% file: tables/ablation-hyper-supp.tex
\begin{center}
	\centering
	\resizebox{0.95\textwidth}{!}{
		\begin{tabular}{|l|l|l||c|c|c|c|c|c|c|}
			\hline
			$K$ & $\alpha_l$ & $\alpha_h$ &\cellcolor{col1}Abs Rel & \cellcolor{col1}Sq Rel & \cellcolor{col1}RMSE  & \cellcolor{col1}RMSE log & \cellcolor{col2}$\delta < 1.25 $ & \cellcolor{col2}$\delta < 1.25^{2}$ & \cellcolor{col2}$\delta < 1.25^{3}$\\
			\hline
			2 & 0.05 & 0.15& 0.055&0.006&0.083&0.075&0.977&0.998&\bf{1.000}\\
			8&0.05&0.15&0.054&0.006&0.084&0.074&0.979&\bf{0.999}&\bf{1.000} \\
			4&0.01&0.3&0.054&0.007&0.087&0.080&0.971&0.997&\bf{1.000} \\
			4&0.05&0.3&0.055&0.007&0.087&0.079&0.976&0.998&\bf{1.000} \\
			4&0.01&0.15&0.053&0.006&0.083&0.075&0.980&0.998&\bf{1.000} \\
			4&0.05&0.15&\bf{0.051}&\bf{0.005}&\bf{0.076}&\bf{0.069}&\bf{0.987}&0.998&\bf{1.000}\\

			\hline
			\end{tabular}}
	\captionof{table}{Hyperparameter analysis. The experiment was conducted on scene0521.
	}

	\label{tab:hyper}
\end{center}

%% file: latex/supp-arxiv.tex
\section{Implementation Details}
 To train the proposed system, we mostly followed NeRF \cite{mildenhall2020nerf}. Specifically, we sampled 64 points in each ray and used a batch of 1024 rays. Since we did not adopt coarse-to-fine strategy in the sampling process, we only need one network (the architecture is same with \cite{mildenhall2020nerf}) to optimize the neural radiance fields. We added random Gaussian noise with zero mean and unit variance to the density $\sigma$ to regularize the network. In addition, following \cite{mildenhall2020nerf}, positional encoding was also employed. Adam was adopted as our optimizer with the initial learning rate as $5 \times 10^{-4}$ and decayed exponentially to $5 \times 10^{-5}$. We utilized PyTorch in our implementation. Each scene was trained with 200K iterations on a single RTX 2080 Ti.

\noindent
\textbf{Error metrics.}
We follow the metrics in \cite{kusupati2020normal,luo2020consistent,murez2020atlas,sinha2020deltas,teed2018deepv2d,zhou2017unsupervised} to evaluate depth estimation results:
\begin{itemize}
\setlength\itemsep{0pt}
\item Abs Rel: $\frac1{|T|}\sum_{y\in T}|y - y^*| / y^*$ 
\item Sq Rel: $\frac1{|T|}\sum_{y\in T}||y - y^*||^2 / y^*$ 
\item RMSE: $\sqrt{\frac1{|T|}\sum_{y\in T}||y - y^*||^2}$ 
\item RMSE log: $\sqrt{\frac1{|T|}\sum_{y\in T}||\log y - \log y^*||^2}$  
\item $\delta < t$: \% of $y$ s.t. $\max(\frac{y}{y^*},\frac{y^*}{y}) = \delta < t$
\end{itemize}
where $y$ and $y^*$ indicate predicted and groundtruth depths respectively, and T indicates all pixels on the depth image. 

\section{Baseline Method Details}
We compared our results with several state-of-the-art depth estimation method, which can be roughly classified as four categories: 

\noindent \textbf{Conventional multi-view stereo:} COLMAP\cite{schonberger2016structure,schonberger2016pixelwise}, ACMP \cite{xu2020planar}. COLMAP is a non-learning MVS method for 3D reconstruction building upon PatchMatch stereo \cite{bleyer2011patchmatch}. Based on COLMAP, ACMP introduces planar models to solve low-textured areas in complex indoor environments. 

\noindent \textbf{Learning-based multi-view stereo:} DELTAS \cite{sinha2020deltas},  Atlas \cite{murez2020atlas}. These two methods are trained on ScanNet with groundtruth depth supervision. For DELTAS, we used two neighboring frames as the reference frames.

\noindent \textbf{Monocular depth estimation:} Mannequin Challenge \cite{li2019learning}. Mannequin Challenge is a state-of-the-art monocular depth estimation method. We directly used their pretrained weight for evaluation. 

\noindent \textbf{Video-based depth estimation:} CVD \cite{luo2020consistent}, DeepV2D \cite{teed2018deepv2d}. For video-based methods, we sorted images in a scene according to the timeline. DeepV2D is trained on ScanNet with groundtruth depth supervision.

\section{Hyperparameter Analysis}
To further demonstrate the effectiveness of our method, we did hyperparameter analysis for the number of used minimum errors $K$, and the bounds $\alpha_l$, $\alpha_h$ used in the guided sampling process. The experiments were conducted on scene0521. Table \ref{tab:hyper} shows experimental results. We find that using a $K$ that is too small or too large will degrade the performance. On the one hand, it is possible to satisfy the multi-view consistency check although the depths are not correct. Small $K$ will increase the probability of this 
phenomenon. On the other hand, there are pixels that do not overlap across some view pairs. Thus, the projection errors on some views are invalid and a large $K$ may cover these invalid views. In addition, a large upper bound $\alpha_h$ or a small lower bound $\alpha_l$ for sampling range will lead to worse results, which indicates the necessity to set bounds in sampling process.